\documentclass{article}

\usepackage{microtype}
\usepackage{graphicx}
\usepackage{subfigure}
\usepackage{booktabs} 
\usepackage{hyperref}

\usepackage[accepted]{icml2024}

\usepackage{amsmath}
\usepackage{amssymb}
\usepackage{mathtools}
\usepackage{amsthm}

\usepackage[capitalize,noabbrev]{cleveref}

\theoremstyle{plain}

\theoremstyle{definition}

\theoremstyle{remark}

\usepackage[textsize=tiny]{todonotes}

\icmltitlerunning{Temporal Graph MLP Mixer for Spatio-Temporal Forecasting}

\begin{document}
\twocolumn[
\icmltitle{Temporal Graph MLP Mixer for Spatio-Temporal Forecasting}



\icmlsetsymbol{equal}{*}

\begin{icmlauthorlist}
\icmlauthor{Muhammad Bilal}{equal,uni}
\icmlauthor{Luis Carretero López}{equal,uni}
\end{icmlauthorlist}

\icmlaffiliation{uni}{Department of Computer Science, ETH Zurich, Zurich, Switzerland}

\icmlcorrespondingauthor{Muhammad Bilal}{mbilal@ethz.ch}
\icmlcorrespondingauthor{Luis Carretero López}{lcarretero@ethz.ch}


\vskip 0.3in
]

\footnotetext[1]{Department of Computer Science, ETH Zurich, Zurich, Switzerland}




\begin{abstract}
Spatiotemporal forecasting is critical in applications such as traffic prediction, climate modeling, and environmental monitoring. However, the prevalence of missing data in real-world sensor networks significantly complicates this task. In this paper, we introduce the Temporal Graph MLP-Mixer (T-GMM), a novel architecture designed to address these challenges. The model combines node-level processing with patch-level subgraph encoding to capture localized spatial dependencies while leveraging a three-dimensional MLP-Mixer to handle temporal, spatial, and feature-based dependencies. Experiments on the AQI, ENGRAD, PV-US and METR-LA datasets demonstrate the model's ability to effectively forecast even in the presence of significant missing data. While not surpassing state-of-the-art models in all scenarios, the T-GMM exhibits strong learning capabilities, particularly in capturing long-range dependencies. These results highlight its potential for robust, scalable spatiotemporal forecasting. 
\end{abstract}

\section{Introduction}
Most timeseries data is collected through sensor networks (SN) over time across different locations. Analyzing and forecasting this data is crucial in a variety of applications such as traffic flow prediction, climate modeling, and environmental monitoring as it helps with decision-making and resource allocation \cite{ghaderi2017deep}. This is a prominent problem in deep learning called spatio-temporal forecasting \cite{ghaderi2017deep}. A key challenge in spatiotemporal forecasting, particularly in real-world SNs, is the prevalence of missing data \cite{marisca2024graph, cini2021filling}. Sensor failures, communication breakdowns, or other operational issues frequently disrupt data collection, resulting in incomplete or irregular sequences \cite{marisca2024graph, cini2021filling}. 

Most existing spatiotemporal models assume the availability of complete and regularly sampled input data, limiting their applicability to these real-world scenarios \cite{marisca2024graph}. Missing data disrupts the spatial and temporal dependencies that these models rely on, leading to a degradation in predictive performance \cite{cini2023graph, cini2023graphb}. When missing data occurs randomly and sporadically, the localized processing capabilities of the models can act as an effective regularization mechanism. Observations close in time and space can provide sufficient information to impute missing values and maintain forecasting accuracy \cite{cini2021filling}. 

However, significant challenges arise when missing data occurs in large, contiguous blocks, such as when sensor failures persist over extended periods or affect entire portions of a network. In these cases, the underlying spatiotemporal dynamics can only be captured by reaching valid observations that are spatially or temporally distant \cite{Marisca2022}. This requires models to expand their receptive fields significantly, incorporating distant yet relevant information. Expanding a model’s receptive field introduces its own set of trade-offs. Deeper layers and more extensive processing can attenuate faster temporal dynamics, making it difficult for models to capture localized patterns in data. \cite{rusch2023survey}.

\subsection{The State of the Field}
Existing approaches to spatiotemporal forecasting can be broadly categorized into multivariate forecasting methods and graph-based models \cite{longa2023graph}. Multivariate forecasting approaches primarily focus on time series patterns, while Spatio-Temporal Graph Neural Networks (STGNN) model spatial and temporal dependencies explicitly. While STGNNs have gained significant attention, their robustness and effectiveness in handling scenarios with missing data—especially in complex cases involving substantial missing data—remains relatively underexplored \cite{marisca2024graph}.

Empirically, GNN-based models have consistently demonstrated superior performance in spatiotemporal forecasting tasks, earning them widespread popularity in the research community \cite{longa2023graph}. However, this trend raises a critical question: why do GNNs perform better, especially when multivariate models sometimes show higher accuracy on similar datasets in specific studies \cite{rnnSurvey, luo2024lsttn, wang2024mixturemodel, cai2020traffictransformer, liu2023vanillatransformer}? Notably, existing studies often fail to investigate or explain the reasons behind this phenomenon, leaving a significant gap in understanding the relative strengths and limitations of these techniques.

\subsection{Contributions}
In this paper, we address these open questions and make the following key contributions:

\begin{enumerate}
    \item \textbf{Understanding Performance Disparities:}
    We provide a comprehensive analysis of the scenarios under which GNN-based models outperform multivariate forecasting models. By examining the interplay between spatial and temporal inductive bias, we identify the conditions where each technique excels, offering a clearer picture of their comparative advantages.
    \item \textbf{Proposing a Novel Temporal Graph-Mixer Model:}
    We introduce Temporal Graph-Mixer (TGM), a novel architecture specifically designed for temporal graph data. This model builds on the Graph ViT/MLP-Mixer \cite{he2023generalization} by integrating a temporal
channel into the model, allowing it to learn temporal dynamics such as trends, periodicities, and localized patterns.
    \item \textbf{Improved Robustness in Missing Data Scenarios:}
    One of the key strengths of the Graph-Mixer model is its ability to model long range dependencies. By being able to model long range dependencies, TGM is able to utilize a larger receptive field to accurately forecast even with large contiguous missing data.  
\end{enumerate}

\section{Models and Methods}

We adopted a first principles approach to formulate a hypothesis to address the problem of missing large contiguous data in the input for spatiotemporal forecasting. This process began with an extensive review of existing literature to understand the efficacy and limitations of various methods. The literature review revealed two predominant approaches: low-inductive-bias models such as transformers and recurrent neural networks (RNNs) and spatiotemporal graph networks \cite{longa2023graph}. The latter are popular because they explicitly encode the spatial and temporal dependencies. Given the nature of our problem, we sought to determine the conditions under which each approach performs optimally and why.

Our hypothesis was that for static graphs, where nodes and edges remain unchanged over time, models with low inductive bias could perform well. This was based on recent findings emphasizing the role of inductive bias in model generalization, particularly when sufficient data and computational resources are available \cite{bachmann2024scaling}. We posited that low-spatial-bias models could generalize effectively in the presence of large missing data and, if successful, could guide the design of a low-inductive-bias architecture tailored to spatiotemporal forecasting under such conditions.

\subsection{Comperative Study}
To test our hypothesis, we conducted experiments using the METR-LA dataset, the largest benchmark dataset for static temporal graphs \cite{li2017diffusion}. The dataset consists of traffic flow data collected by 207 sensors in the Los Angeles County highway system, aggregated in five-minute intervals, spanning March to June 2012. We split the data into training (70\%), validation (10\%), and test (20\%) sets, evaluating model performance using the Mean Absolute Error (MAE) metric on 12-step-ahead predictions. Our proposed low-spatial-inductive-bias model consisted of a five-layered Long Short-Term Memory (LSTM) network, followed by two fully connected layers. The model was trained for 500 epochs with batch normalization, dropout, and regularization strategies to prevent over fitting. Dropout rates were incrementally reduced from 0.8 to 0.4 across successive layers.

For baseline comparisons, we implemented three top performing models from the literature: the Diffusion Convolutional Recurrent Neural Network (DCRNN), which integrates graph-based diffusion convolution with recurrent neural networks to capture spatial and temporal dependencies \cite{li2017diffusion}; Graph WaveNet, a graph neural network utilizing adaptive graph convolutions and dilated causal convolutions for efficient modeling of spatial relationships and long-term temporal dependencies \cite{waveNet2019graph}; and the Graph Multi-Attention Network (GMAN), a transformer-based model leveraging graph attention mechanisms to capture spatiotemporal correlations at multiple scales \cite{zheng2020gman}. These models represent the state-of-the-art in spatiotemporal forecasting and served as benchmarks to evaluate our approach.

\subsection{Temporal Graph MLP mixer}
Forecasting with missing data requires models to expand their receptive fields significantly. However, expanding the receptive field in Graph Neural Networks (GNNs) introduces two major challenges: modeling long-range dependencies and mitigating over-smoothing. One effective method to deal with this problem is presented in the foundational missing data paper \cite{marisca2024graph} which is to compute separate spatial and temporal representations at multiple scales. These representations are subsequently combined using a soft attention mechanism, enabling the model to focus selectively on relevant features across scales. Graph MLP-Mixer demonstrates excellent capabilities in modeling long-range dependencies \cite{he2023generalization}. Incorporating our findings from our comparative analysis, we hypothesized that a temporal Graph MLP would be particularly well-suited for spatiotemporal forecasting tasks involving significant missing data. 

The Temporal Graph MLP Mixer model comprises three main components: an encoder, an MLP-Mixer core, and a readout mechanism. The node encoder processes timeseries data and boolean validity masks - for imputed or real data - into latent representations, which are handled by a node-level MLP-Mixer. The graph is divided into subgraphs using the METIS algorithm, with one-hop neighborhoods added to ensure edge coverage. A shallow GNN encodes these patches, followed by mean-pooling to create compact, localized representations which are forwarded to the MLP-Mixer core. The MLP-Mixer core mixes across spatial, feature, and temporal dimensions to capture complex dependencies. The readout mechanism aggregates outputs from the patch MLP-Mixer, averaging representations for nodes in multiple patches. This information is concatenated with node-level outputs and passed through a final MLP (similar to the temporal projection in \cite{chen2023tsmixer}) to produce temporal forecasts. 

We conducted experiments on the AQI, ENGRAD, PV-US and METR-LA datasets following the methodology in \cite{marisca2024graph}. This involved sliding windows and horizons over the dataset for temporal predictions. Both datasets inherently contained some missing data, which were augmented with additional synthetic missing data patterns (Point, BlockT, and BlockST)\cite{marisca2024graph}. For all missing data — both real and synthetic — we employed a simple imputation strategy, filling gaps with the last observed value at the corresponding node. 

The model predicted the next horizon of timesteps for each node, but loss and evaluation metrics during training were computed on non-missing data. This was to avoid biasing the model towards only predicting the last observed value, which could trivially perform well in datasets with a high proportion of missing data.  During testing, metrics were computed on all data, including synthetically masked data, ensuring that the model was evaluated on realistic fault scenarios. All datasets were split into 70\%/10\%/20\% for training, validation, and testing. We trained the models using AdamW optimizer with learning rate scheduling, weight decay, dropout, and early stopping for regularization and stability. The results of these experiments are detailed in the Results section.

\section{Results}

\subsection{Comparative Study}
In our comparative study, we evaluated our LSTM model against three implemented baselines (DCRNN, Graph WaveNet, and GMAN) and three additional models from literature (TITAN \cite{wang2024mixturemodel}, STAEformer \cite{liu2023vanillatransformer}, and Traffic Transformer \cite{cai2020traffictransformer}). The results are summarized in Table 1.

Notably, the LSTM model demonstrated competitive performance relative to state of the art (SOTA) models. Figure 1 illustrates the training dynamics, showing that the LSTM’s Mean Absolute Error (MAE) decreases logarithmically over epochs, requiring a substantial number of epochs to converge. While further reductions in error may have been achievable with extended training, the current results were sufficient to validate our hypothesis regarding the potential of low-inductive-bias models in spatiotemporal forecasting tasks.

\begin{table}[!ht]
\centering
\begin{tabular}{@{}llcc@{}}
\toprule
\textbf{Data}    & \textbf{Models}        & \textbf{MAE} & \textbf{MAPE} \\ \midrule
METR-LA          & \textbf{FC-LSTM}      & \textbf{3.52} & \textbf{10.14} \\
                 & DCRNN                 & 3.56          & 10.35         \\
                 & Graph WaveNet         & 3.51          & 10.06         \\
                 & GMAN                  & 3.56          & 10.33         \\
                 & TITAN                 & 3.08          & 8.43          \\
                 & STAEformer            & 3.34          & 9.70          \\
                 & Traffic Transformer   & 3.28          & 9.08          \\
            & \textbf{T-GMM} & \textbf{4.22}  & \textbf{10.83} \\
                 \bottomrule
\end{tabular}
\caption{Comparison of MAE and MAPE for different models on the METR-LA dataset.}
\label{tab:metr_la_metrics}
\end{table}

\subsection{Temporal Graph MLP Mixer}
The performance of the T-GMM on the METR-LA dataset highlights its potential in capturing spatiotemporal patterns, despite not achieving state-of-the-art results. On this dataset, the model achieved a Mean Absolute Error (MAE) of 4.22 and a Mean Absolute Percentage Error (MAPE) of 10.83. While these metrics fall short of the best-performing models listed in Table 1, they are a strong indication that the T-GMM is effectively learning. As illustrated in Figure 1, the model demonstrates the ability to capture meaningful patterns in its 12-step-ahead predictions, providing forecasts that closely follow the trends in the true data. These results underscore the model's viability as a promising architecture for spatiotemporal forecasting tasks.

On the HDTTS datasets, our model showed mixed performance, as shown in Table 2. For GraphMSO with Point missing data, T-GMM achieved strong results that outperformed most baseline models with an MAE of 0.078. However, performance degraded on BlockT and BlockST patterns (MAE 0.470 and 1.192 respectively), though remaining competitive but not reaching state-of-the-art levels. The AQI dataset, which inherently contains 25\% missing data, proved more challenging - T-GMM's performance was notably worse compared to GraphMSO across all metrics, with MAEs of 21.13 and 22.06 for original and Point patterns respectively.

\begin{figure}[!ht]
    \centering
    \includegraphics[width=\linewidth]{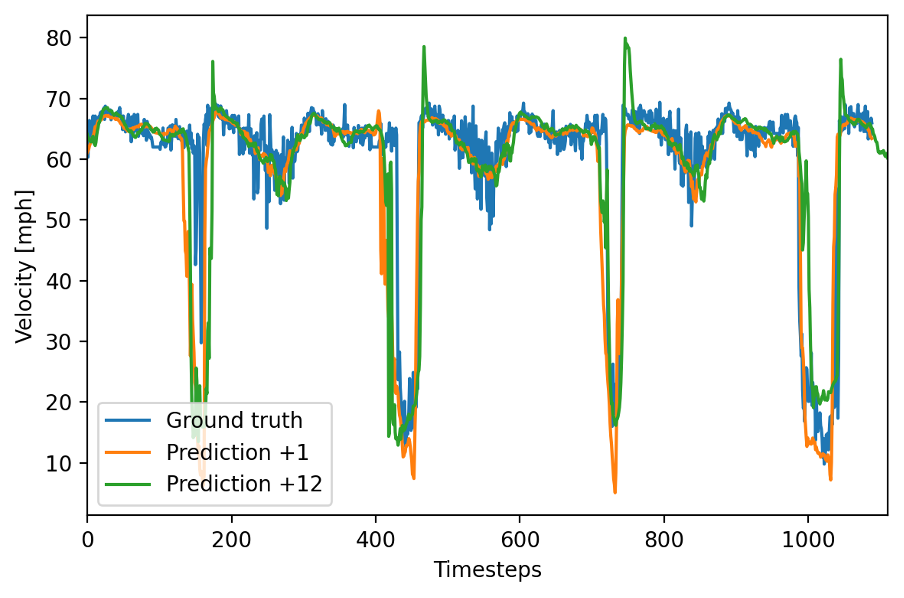}
    \caption{Prediction of the T-GMM on the METR-LA dataset.}
    \label{fig:metr_la_pred}
\end{figure}

\begin{table}[!ht]
\centering
\begin{tabular}{@{}llccc|cc@{}}
\toprule
\textbf{Models}        & \multicolumn{3}{c|}{\textbf{GraphMSO}} & \multicolumn{2}{c}{\textbf{AQI}} \\ 
\cmidrule(lr){2-4} \cmidrule(lr){5-6}
                       & \textbf{Point} & \textbf{BlockT} & \textbf{BlockST} & \textbf{Orig.} & \textbf{Point} \\ \midrule
GRU          & 0.346          & 0.639             & 1.137              & 18.17             & 19.19          \\
DCRNN                  & 0.291          & 0.645             & 1.103              & 16.99             & 17.51          \\
AGCRN                  & 0.067          & 0.366             & 1.056              & 17.19             & 17.92          \\
GWNet                  & 0.089          & 0.340             & 0.955              & 15.89             & 16.39          \\
T\&S-IMP               & 0.118          & 0.323             & 0.935              & 16.54             & 17.13          \\
T\&S-AMP               & 0.063          & 0.293             & 0.868              & 16.15             & 16.58          \\
TTS-IMP                & 0.113          & 0.271             & 0.697              & 16.25             & 16.90          \\
TTS-AMP                & 0.096          & 0.251             & 0.669              & 15.63             & 16.15          \\
HDTTS1             & 0.058          & 0.247             & 0.651              & 15.50             & 15.94          \\
HDTTS2             & 0.062          & 0.261             & 0.679              & 15.35             & 15.76          \\
\textbf{T-GMM}         & \textbf{0.078} & \textbf{0.470}    & \textbf{1.192}     & \textbf{21.13}    & \textbf{22.06} \\
\bottomrule
\end{tabular}
\caption{Comparison of metrics for different models on the GraphMSO and AQI datasets.}
\label{tab:grouped_metrics}
\end{table}

\section{Discussion}

\subsection{Comparative Study}

Our results demonstrate that the low spatial inductive bias LSTM model achieved competitive performance compared to SOTA models, supporting our hypothesis that such architectures can effectively learn complex spatiotemporal patterns without strong built-in spatial biases. The only requirement is having a large dataset and sufficient compute. This result aligns with the scaling laws suggesting that the trade-off for increased generalizability is primarily in training time and data \cite{bachmann2024scaling}. Although our experiments confirmed the hypothesis that low-inductive-bias models can generalize effectively, further experimentation to fine-tune hyper-parameters could explore whether accuracy gains are sufficient to surpass SOTA models. Such investigations would provide additional evidence to support the hypothesis that low-inductive-bias models achieve superior generalization with sufficient training resources and data.

The SOTA models from the literature, such as TITAN, STAEformer, and Traffic Transformer, achieve higher accuracy partly due to their tailored architectures that have been extensively hyperparameter-tuned for optimal performance. In contrast, our focus was on demonstrating that an FC-LSTM model can achieve competitive accuracy when provided with sufficient data and training time, even without hyperparameter fine-tuning.

\begin{figure}[!ht]
    \centering
    \includegraphics[width=\linewidth]{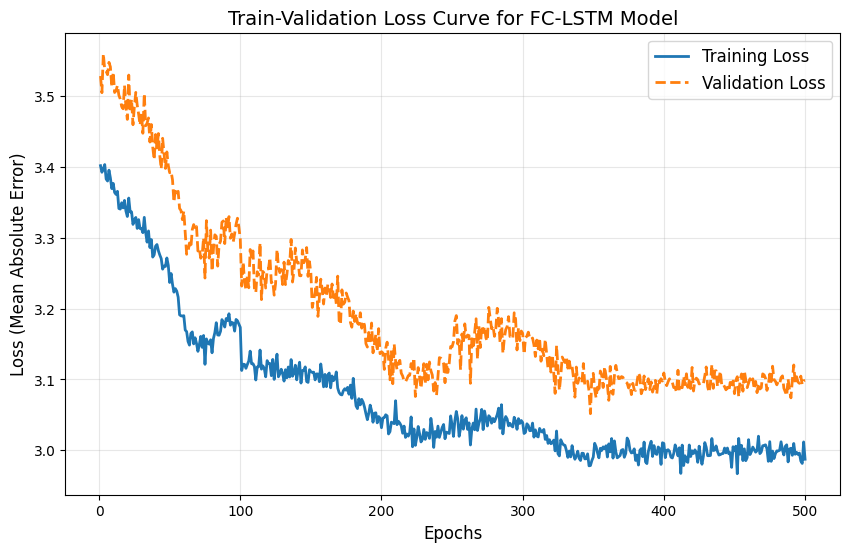}
    \caption{Train-validation loss curve for the FC-LSTM.}
    \label{fig:loss_curve}
\end{figure}

One significant drawback of the low-inductive-bias method is its requirement for a large number of epochs to achieve high accuracy (see Figure 1). In contrast, state-of-the-art (SOTA) models demonstrate faster convergence. While it can be argued that this reliance on strong biases may limit their generalization capabilities when encountering novel or unseen patterns, this aspect was not explored in our comparative study. However, we address this in the second section of our research, where we introduce a novel architecture, the Temporal Graph MLP Mixer, and evaluate its forecasting performance on datasets with missing values, comparing it to SOTA models.

\subsection{Temporal Graph MLP mixer}

A key challenge in applying MLP-Mixer architectures to temporal forecasting is their tendency to overfit, particularly with non-stationary real-world data. While common practice uses sliding windows and chronological train/validation/test splits (70/10/20), this assumes stationarity. We observed significant non-stationarity in the MetrLA dataset, with naturally missing data varying from 9\% in training to 100\% in validation periods. This poses challenges for MLP-Mixer models generalizing across different distributions. Our model performed well on the stationary synthetic GraphMSO dataset, achieving similar train and validation metrics due to its simple harmonic patterns. However, performance degraded when we introduced non-stationarity through BlockT and BlockST patterns, highlighting generalization challenges with non-stationary data.

A key limitation of the current GMM is high memory usage. The patching process assigns each node to multiple patches (typically 3-10) due to one-hop neighborhood expansion after subgraph partitioning. During GNN patch-encoding, this results in 3-10x more node computations, since the GNN must process every node in every patch. While manageable for small (spatially) datasets like MetrLA and GraphMSO, this leads to prohibitive memory requirements for larger datasets like PV-US (1000+ nodes). The original Graph-MLP-Mixer paper avoided this issue by only using datasets with at most 150 average nodes.

Our analysis of inductive biases in GNNs reveals two key properties: distance-based information exchange between connected nodes, and invariance to node indexing. While GNNs like GCN and GAT are designed for both properties, our static graphs with dynamic node features only require the first. Since the graph topology is fixed, node permutation invariance is unnecessary. The original Graph-MLP-Mixer maintains this invariance through permutation-invariant encodings, but for our use case we could simplify by removing these components. In the future, we could thus replace the GNN encoder with the original MLP-Mixer's encoder, reducing spatial bias to just the patching process grouping nearby nodes. This would also create a more lightweight model and resolve the memory scaling issues mentioned above.

\section{Summary}

This study introduced the Temporal Graph MLP Mixer, a novel architecture for spatiotemporal forecasting that addresses the challenges posed by incomplete and irregular data. Building upon the Graph MLP Mixer, the model captures complex dependencies across spatial, temporal, and feature dimensions. Through experiments on the METR-LA and HDTTS datasets, we demonstrated the model’s capability to generalize across diverse scenarios, achieving competitive results and effectively handling long-range dependencies, even with significant missing data. The comparative study provided insights into the strengths and weaknesses of low-inductive-bias models relative to state-of-the-art models. While SOTA models excelled in accuracy and convergence speed, our findings highlight the potential of low-inductive-bias models to achieve comparable performance with sufficient data and training. These models also offer greater flexibility for generalization to novel or unseen patterns, albeit at the cost of increased training time. Overall, this study demonstrates the utility of the Temporal Graph MLP Mixer and contributes to a deeper understanding of the trade-offs in spatiotemporal forecasting model design.
\section{Reproducibility statement}
All the code, data and instructions to reproduce the experiments in this paper can be found on the following Github Repository: https://github.com/Mu-Bilal/Temporal-Graph-MLP-Mixer

\nocite{}

\bibliography{proposal.bib}
\bibliographystyle{icml2024}

\newpage
\appendix
\onecolumn
\section{Experimental settings}
\subsection{Software}
Our experiments were conducted using PyTorch Lightning as the main deep learning framework, with PyTorch as the underlying engine. For graph neural network operations, we utilized PyTorch Geometric (PyG). Experiment tracking and model checkpointing were handled through Weights \& Biases (wandb). The codebase was developed in Python 3.10.

\subsection{Graph MLP-Mixer Training settings}
For training, we used a consistent model architecture across all datasets, with the following key components:

The model uses a hierarchical architecture with:
\begin{itemize}
    \item GINEConv \cite{hu2019strategies} as the graph neural network layer type
    \item 2 GNN layers for local message passing
    \item 2-3 patch mixer layers and 3-4 node mixer layers for global information exchange
    \item Single layer MLP readout for final predictions
\end{itemize}

Key hyperparameters were slightly adjusted per dataset but generally followed:
\begin{itemize}
    \item Learning rate of 1e-3 with a decay factor of 0.5 and minimum learning rate of 1e-5
    \item Early stopping patience of 5 epochs monitoring validation loss
    \item Dropout rates between 0.1-0.8 for different components (GNN, mixer layers, readout)
    \item Feature dimensions of 64-128 for both patch and node representations
\end{itemize}

Training was performed with gradient accumulation and batch sizes optimized per dataset.

\end{document}